\documentclass{article}
\usepackage{graphicx}
\usepackage{amsmath,amsthm,amsfonts}
\newcommand{\defeq}{\mathrel{\mathop:}=}

\newcommand{\Ex}{\mathbb{E}}
\newcommand{\real}{\mathbb{R}}

\newcommand{\muHalf}{{\mu^{\frac{1}{2}}}}

\usepackage{times}
\usepackage{graphicx} 
\usepackage{subfigure} 

\usepackage{natbib}

\usepackage{algorithm}
\usepackage{algorithmic}

\usepackage{hyperref}

\usepackage[accepted]{icml2015}

\icmltitlerunning{A Linear Dynamical System Model for Text}

\begin{document} 

\twocolumn[
\icmltitle{A Linear Dynamical System Model for Text}

\icmlauthor{David Belanger}{belanger@cs.umass.edu}
\icmladdress{College of Information and Computer Sciences, University of Massachusetts Amherst}
\icmlauthor{Sham Kakade}{skakade@microsoft.com}
\icmladdress{Microsoft Research}

\icmlkeywords{boring formatting information, machine learning, ICML}

\vskip 0.3in
]

\begin{abstract}
Low dimensional representations of words allow accurate NLP models to be trained on limited annotated data. While most representations ignore words' local context, a natural way to induce context-dependent representations is to perform inference in a probabilistic latent-variable sequence model. Given the recent success of continuous vector space word representations, we provide such an inference procedure for continuous states, where words'  representations are given by the posterior mean of a linear dynamical system. Here,  efficient inference can be performed using Kalman filtering. Our learning algorithm is extremely scalable, operating on simple cooccurrence counts for both parameter initialization using the method of moments and subsequent iterations of EM. In our experiments, we employ our inferred word embeddings as features in standard tagging tasks, obtaining significant accuracy improvements. Finally, the Kalman filter updates can be seen as a linear recurrent neural network. We demonstrate that using the parameters of our model to initialize a non-linear recurrent neural network language model reduces its training time by a day and yields lower perplexity. 
\end{abstract}

\section{Introduction}
In many NLP applications, there is limited available labeled training data, but tremendous quantities of unlabeled, in-domain text. An effective semi-supervised learning technique is to learn \textit{word embeddings} on the unlabeled data, which map every word to a low dimensional dense vector~\citep{bengio2006neural,mikolov2013efficient,pennington2014glove}, and then use these as features for supervised training on the labeled data~\citep{turian2010word,passos-kumar-mccallum:2014:W14-16,bansal2014tailoring}. Furthermore in many deep architectures for NLP, the first layer maps words to low-dimensional vectors, and these parameters are initialized with unsupervised embeddings~\citep{collobert2011natural,socher2013reasoning,vinyals2014grammar}.

Most of these methods embed~\textit{word types}, i.e., words independent of local context, as opposed to \textit{work tokens}, i.e., instances of words within their context. Ideally we would have a different representation per token.  For example, depending on the context, ``bank" is  the side of a river or a financial institution. Furthemore, we would like such embeddings to come from a probablistic sequence model that allows us to study the transition dynamics of text generation in low dimensional space. 

We present a method for obtaining such context-dependent token embeddings, using a generative model with a vector-valued latent variable per token and performing posterior inference for each sentence. Specifically, we employ a Gaussian \textit{linear dynamical system} (LDS), with efficient inference from a \textit{Kalman filter}. To learn the LDS parameters, we use a two-stage procedure, initializing with the method of moments, and then performing EM with the approximate second order statistics (ASOS) technique of~\citet{martens2010learning}. Overall, after taking a single pass over the training corpus, the runtime of our approximate maximum-likelihood estimation (MLE) procedure is independent of the amount of training data since it operates on aggregate co-occurrence counts. Furthermore, performing inference to obtain token embeddings has the same time complexity as widely-used discrete first-order sequence models.

We fit the LDS to a one-hot encoding of each token in the input text sequence. Therefore, the LDS is a mis-specified generative model, since draws from it are not proper indicator vectors.  However, we embrace this multivariate Gaussian model instead of a continuous-state dynamical system with a multinomial link function because the Gaussian LDS offers several desirable scalability properties: (1) Kalman filter inference is simple and efficient (2) using ASOS, the cost of our learning iterations does not scale with the corpus size, (3) we can initialize EM using a method-of-moments estimator that requires a single SVD of a co-occurrence matrix,  (4) our M-step updates are simple least-squares problems, solvable in closed form, (4) if we had used a multinomial link function, we would have performed inference using \textit{extended Kalman filtering}, which makes a second-order approximation of the log-likelihood, and thus leads to a Gaussian LDS anyway~\citep{ghahramani1999learning}, and (5) by using EM, we avoid stochastic-gradient-based optimization, which requires careful tuning for nonconvex problems. A naive application of our method scales to large amounts of training data, but not high-dimensional observations. In response, the paper contributes a variety of novel methods for scaling up our learning techniques to handle large input vocabularies. 

We employ our inferred token embeddings as features for part of speech (POS)  and named entity recognition (NER) taggers. For POS, we obtain a 30\% relative error reduction when applying a local classifier to our context-dependent embeddings rather than Word2Vec context-independent embeddings~\citep{mikolov2013efficient}. When using our token embeddings as additional features in lexicalized POS and NER taggers, which already have explicit features and test-time inference for context-dependence, we obtain signficant gains over the baseline, performing as well as using Word2Vec embeddings. We also present experiments demonstrating that the transition dynamics of the LDS capture salient patterns, such as transforming first names into last names.

Finally, the functional form of the Kalman filter update equations for our LDS are identical to the updates of a recurrent neural network (RNN) language model without non-linearities~\citep{mikolov2012statistical}. A key difference between the LDS and an RNN, however, is that the LDS provides a natural backwards pass, using \textit{Kalman smoothing}, where a token's embedding depends on text to both the right and left. Drawing on the parallelism between filtering and the RNN, we use the LDS parameters, which can be estimated very quicky using our techniques, to initialize gradient-based optimization of a nonlinear RNN. This yields a signficant decrease in perplexity vs. the baseline RNN, and only requires 70\% as many training epochs, saving 1 day on a single CPU core. 

\section{Related Work}
\label{sec:related-work}
We provide a continuous analog of popular discrete-state generative models used in NLP for inducing class membership for tokens, including class-based language models~\citep{brown1992class,chelba2000structured} and induction of POS tags~\citep{christodoulopoulos2010two}. In particular, Brown clusters~\citep{brown1992class} are commonly used by practioners with lots of unlabeled in-domain data. 

Our learning algorithm is very scalable because it operates on aggregate count matrices, rather than individual tokens. Similar algorithms have been proposed for obtaining type-level embeddings via matrix factorization~\citep{pennington2014glove,levy2014neural}. However, these are context-independent and ignore the transition dynamics that link tokens' embeddings. Furthermore, they require careful tuning of stochastic gradient methods. Previous methods for token-level embeddings either use a rigid set of prototypes~\citep{huang2012improving,neelakantan2014efficient} or 
embed the token's context, ignoring the token itself~\citep{dhillon11multiviewcca}.

For learning discrete-state latent variable models, spectral learning methods also use count matrices, and thus are similarly scalable~\citep{anandkumar2014tensor}. However, an LDS offers key advantages: we do not use third-order moments, which are difficult to estimate, and we perform approximate MLE, rather than the method of moments, which exhibits poor statistical efficiency. 

Recently, RNNs have been used to provide impressive results in NLP  applications including translation~\citep{sutskever2014sequence}, language modeling~\citep{mikolov2014learning}, and parsing~\citep{vinyals2014grammar}. We do not attempt to replace these with a Kalman filter, as we expect non-linearities are crucial for capturing long-term interactions and rigid, combinatorial constraints in the outputs. However, RNNs training can take days, even on GPUs, and requires careful tuning of stochastic gradient step sizes. Given the scalability of our parameter-free training algorithm, and our favorable preliminary results using the LDS to initialize a nonlinear RNN, we encourage further work on  using linear latent variable models and the Gaussian approximations of multinomial data to develop sophisticated initialization methods. Already, practitioners have started using such techniques for initializing simple nonlinear deep neural networks using the recommendations of~\citet{saxe2013exact}. Finally, our work differs from~\citet{pasa2014pre}, who initialize an RNN using spectral techniques, in that we perform maximum-likelihood learning. We found this crucial for good performance in our NLP experiments. 

\section{Background: Gaussian Linear Dynamical Systems}

We consider sequences of observations  $w_1, \ldots, w_n$, where each
$w_i$ is a $V$-dimensional vector. A Gaussian LDS follows the following generative model~\citep{kalman,roweis1999unifying}: 
\begin{eqnarray}
x_t &= A x_{t-1} + \eta\\
w_t &= C x_{t} + \epsilon, \label{eq:gen2}
\end{eqnarray}

where $h < V$ is the dimensionality of the hidden states $x_t$ and
$\epsilon \sim N(0,D)$, $\eta \sim N(0,Q)$.  For simplicity, we assume $x_0$ is constant. 

The latent space for $x$ is completely unobserved and we could choose
any coordinate system for it while maintaining the same likelihood value. Therefore, without loss of generality,
we can either fix $A = I$ or $Q = I$, and we fix $Q$. Furthermore, note that the
magnitude of the maximum eigenvalue of $A$ must be no larger than 1 if the system is stable. We assume that the data
we fit to has been centered, in which case the maximum eigenvalue is
strictly less than 1, since this implies $x_t$ is asymptotically mean zero
(independent of $x_0$), so that $x_t$ is also asymptotically mean zero.

Finally, define the covariance at lag $k$ to be 
\begin{equation}
\Psi_k = \Ex[w_{t+k} w_t^\top],
\end{equation} 
which is valid because we assume the data to be mean zero. Our learning algorithms require only a few $\Psi_k$ (up to about $k = 10$ in practice) as input. These matrices can be gathered using a single pass over the data, and their size does not depend on the amount of data. Furthermore, constructing these matrices can be accelerated by splitting the data into chunks, and aggregating separate matrices afterwards.

\subsection{Inference}
The $x_t$ are distributed as a multivariate Gaussian under both the LDS prior and posterior (conditional on observations $w$), so they can be fully characterized by a mean and variance. We use $\hat{x}_t$ and $S_t$ for the mean and covariance under the posterior for $x_t$ given $w_{1:(t-1)}$, computed using \textit{Kalman filtering}, and $\bar{x}_t$ and $S_T$ when considering the posterior for $x_t$ given all the data $w_{1:T}$, computed using \textit{Kalman smoothing}. In Appendix~\ref{app:filter} we provide the full filtering and smoothing updates, which compute different means and variances for every timestep. Note that the updates require inverting a $V \times V$ matrix in every step. 

We employ the widely-used 'steady-state' approximation, which yields substantially more efficient filtering and smoothing updates~\citep{rugh1996linear}.  A key property of filtering and smoothing is that the updates to  $S_t$ and $S_T$ do not depend on the actual observations, but only on the model's parameters. Furthermore, they will converge quickly to time-independent `steady-state' values. Define $\Sigma_1 = \Ex[\hat{x}_t\hat{x}_t^\top| w_{1:(t-1)}]$ to be the aymptotic limit of the covariance $S_t$ under the posterior for each $x_t$ given its history (at steady state, this is shared for all $t$). Here, expectation is taken with respect to both time and the posterior for the latent variables. This satisfies $$\Sigma_1 = A \Sigma_1 A^\top + Q,$$ which can be solved for quickly using fixed point iteration. Similarly, we can solve for $\Sigma_0 = \Ex[\hat{x}_t\hat{x}_t^\top| w_{1:t}]$. Note that steady state, a property of the posterior, is unrelated to the stationary distribution of the LDS, which is unconditional on observations.

Under, the steady state assumption, we can perform filtering and smoothing using substantially more efficient updates. We have:
\begin{align}
\hat{x}_t &= (A - KC A)\hat{x}_{t-1} + K w_t \label{eq:kf1} \\
\bar{x}_{t} &= J \bar{x}_{t+1} + (I - JA) \hat{x}_t\label{eq:kf2} 
\end{align}

Here, the steady-state Kalman gain matrix is:
\begin{equation}
K = \Sigma_1 C^\top S_{ss}^{-1} \in \real^{h \times V},~\label{eq-ss-gain}
\end{equation}
where we define 
\begin{equation}
S_{ss} = C \Sigma_1  C^\top + D,
\end{equation}
the unconditional prior covariance for $w$ under the model.  Note that $(A - KC A) $ is data-independent and can be precomputed, as can the smoothing matrix $J = \Sigma_0 A^\top (\Sigma_1)^{-1}$. For long sequences, steady-state filtering provides asymptotically exact inference. However, for short sequences it is an approximation.

\subsection{Learning: Expectation-Maximization}
See~\citet{ghahramani1996parameter} for a full exposition on learning
the parameters of an LDS using EM. Under the steady-state assumption, the M step requires:
\begin{align}
\label{eq:expectations}
\Ex[ \bar{x}_t \bar{x}_t^\top],\; \Ex[ \bar{x}_t \bar{x}_{t+1}^\top], \; \Ex[ \bar{x}_t w_t^\top],
\end{align}
where the expectation is taken with respect to time and the posterior for the
latent variables. This can be computed using Kalman smoothing and then
averaging over time. The M step can then be done in closed form, since it is
solving least-squares regressions for  $x_{t+1}$ against $x_t$ and $w_t$ against $x_t$ to obtain $A$ and $C$. Lastly,  $D$ can be recovered using:
\begin{align}
\label{eq:D-EM}
D&=\Psi_0- C \Ex \left[\bar{x}_t w_t^\top \right] \nonumber \\
&-  \Ex \left[w_t \bar{x}_t^\top \right] C^\top + C \Ex\left[ \bar{x} \bar{x}_t^\top \right] C^\top
\end{align}

\subsection{Learning: EM with ASOS~\citep{martens2010learning}}

EM requires recomputing the second order
statistics~\eqref{eq:expectations} in every iteration. While these can be computed using Kalman smoothing on the entire training set, we are interested in datasets with billions of timesteps. Fortunately, we can avoid smoothing by employing the ASOS (approximate second order statistics) method of~\citet{martens2010learning}, which directly performs inference about the time-averaged second order statistics. 

Under the steady-state assumption, this is doable because we can recursively define relationships between second order statistics at lag $k$ and at lag $k+1$ using the recursive relationships of the underlying dynamical system. Namely, rather than performing posterior inference by recursively applying the linear operations~\eqref{eq:kf1} and~\eqref{eq:kf2}, and then averaging over time, we switch the order of these operations and apply the linear operators to time-averaged second order statistics. For example, the following equality is an immediate consequences of the filtering equation~\eqref{eq:kf1} (where expectation is with respect to $t$ and the posterior for $x$):
\begin{align}
\Ex [\hat{x}^{t}_t w_t^\top] &= (A - KCA)\Ex[\hat{x}^{t-1}_{t-1} w_t^\top]  + K \Ex[w_t w_t^\top]
\end{align}

ASOS uses a number of such recursions, along with methods for estimating covariances at a time horizon $r$. These covariances can be approximated by assuming that they are exactly described by the current estimate of the model parameters. Therefore,  unlike standard EM, performing EM with ASOS allows us to precompute an empirical estimate of the $\Psi_k$ at various lags (up to about $ r = 10$) and then never touch the data again. Furthermore,~\citep{martens2010learning} demonstrates that the ASOS approximation is consistent. Namely, the error in approximating the time-averaged second order statistics vanishes with infinite data when evaluated at the MLE parameters.  Overall, ASOS scales linearly with $r$ and the cost of multiplying by the $\Psi_k$. 

\subsection{Learning: Subspace Identification}
\label{sec:ssid}
We initialize EM using Subspace Identification (SSID), a family of method-of-moments estimators
that use spectral decomposition to recover LDS parameters~\citep{van1996subspace}.  The rationale for such a combination is that the method of moments is statistically consistent, so performing it on reasonably-sized datasets will yield parameters in the neighborhood of the global optimum, and then EM will perform local hill climbing to find a local optimum of the marginal likelihood. For LDS, this combination yields empirical accuracy gains in~\citet{smith2000}. A related two-stage estimator, where the local search of EM is replaced with a single Newton step on the local likelihood surface, is known to be minimax optimal, under certain local asymptotic normality conditions~\citep{le1974notes}. 

For our particular application, we use SSID as an approximate method, where it is not statistically consistent, due to the mis-specification of fitting indicator-vector data as a multivariate Gaussian. In our experiments, we discuss the superiority of SSID+EM rather than just SSID.  We do not present results using EM initialized randomly rather than with SSID, since we found it very difficult for our high dimensional problems to generate initial parameters that allowed EM to reach high likelihoods. 

We employ the `n4sid' algorithm of~\citet{van1994n4sid}. Define $r$ to be a small integer. Define the $(rV) \times h$ matrix $\Gamma_r = \left[C \; ; \; CA \; ; \; CA^2 \; ; \; \ldots \; ; \; CA^{r-1} \right]$, where `;' denotes vertical concatenation. Also define the x-w covariance at lag 0: $G = \Ex [x_{t+1} w_t^\top]$ and the $ h \times (rV) $  matrix  $\Delta_r = \left[ A^{r-1} G \; A^{r-2}G \ldots AG \; G \right]$. 

%
%

Next, define the Hankel matrix
\begin{equation}
H_r =  \left( \begin{array}{ccccc}
\Psi_r & \Psi_{r-1} & \Psi_{r-2} & \ldots & \Psi_1 \\
\Psi_{r+1} & \Psi_{r} & \Psi_{r-1} & \ldots & \Psi_2 \\
\ldots &&&& \\
\Psi_{2r-1} & \Psi_{2r-2} & \Psi_{r-3} & \ldots & \Psi_{r}
\end{array} \right). \label{eq:hankel-matrix}
\end{equation}

Then, we have $H_r = \Gamma_r \Delta_r$. 

Let $(U,S,V)$ result from a rank-$h$ SVD of an empirical
estimate of $H_r$, from which we set $\Gamma_r = U S^{\frac{1}{2}}$ and
$\Delta_r = S^{\frac{1}{2}} V^\top$. To recover the LDS parameters, we first define $\Delta_r^{1:(r-1)}$ to be the
submatrix of $\Delta_r$ corresponding to the first $(r-1)$
blocks. Similarly, define $\Delta_r^{2:r}$. From the definition of
$\Delta_r$, we have that $A \Delta_r^{2:r} = \Delta_r^{1:(r-1)}$, so
we can estimate $A$ as $A = \Delta_r^{1:(r-1)}
(\Delta_r^{2:r} )^+$. 
Next,  one  can read off an estimate for $C$ as the first block of $\Gamma_r$. Alternatively, since the previous step gives us a value for $A$ one can set up a regression problem similar to the previous step to solve for $C$ by invoking the block structure in $\Gamma_r$. 

Finally, we need to recover the covariance matrix $D$.  We  first find the asymptotic latent covariance $\Sigma_1$ 
using fixed-point iteration $\Sigma_1 = A \Sigma_{1}A^\top + Q$. From this, we set $D$ using a similar update as~\eqref{eq:D-SSID}, which uses statistics of the LDS posterior, except here $\Sigma_1$ is unconditional on data and is purely a function of the LDS parameters. 
\begin{equation}
D = \Psi_0    - C \Sigma_{1}C^\top. \label{eq:D-SSID}
\end{equation}

\section{Linear Dynamical Systems for Text}
\label{sec:text-lds}

\begin{algorithm}[tb]
   \caption{Learning an LDS for Text}
   \label{alg:generic}
\begin{algorithmic}
   \STATE {\bfseries Input:} \\ 
\STATE  Text Corpus, approximation horizon $r$ (e.g., 10)
   \STATE {\bfseries Output:} \\
\STATE LDS parameters and filtering matrices: $(A,C,D,K,J)$
\STATE
\STATE Gather the matrices $\Psi_k = \Ex[w_{t+k} w_t^\top]$ $(k < r)$ 
\STATE  $W \leftarrow \Psi^{-\frac{1}{2}}_0$  (diagonal whitening matrix)
\STATE $\Psi_k \leftarrow W \Psi_k W^\top$ (whitening)
\STATE $\text{Params} \leftarrow \text{Subspace\_ID}(\Psi_0, \ldots, \Psi_{r})$
\STATE $\text{Params} \leftarrow$ ASOS\_EM(Params,$\Psi_0, \ldots, \Psi_{r}$)


\end{algorithmic}
\end{algorithm}

We fit an LDS to text using SSID to initialize EM, where the E step is performed using ASOS. A summary of the procedure is provided in Algorithm~\ref{alg:generic}. SSID and ASOS scale to extremely large training sets, since they only require the $\Psi_k$ matrices, for small $k$. However, they can not directly handle the very high dimensionality of text observations (vocabulary size $V \approx 10^5$). In this section, we first describe particular properties of the data distribution. Then, we describe novel techniques for leveraging these properties to yield scalable learning algorithms. 

Define $\tilde{w}_t$ as an indicator vector that is 1 in the index of the word at time $t$ and define  $\mu_i$ to be the corpus frequency of word type $i$. We fit to the mean-zero observations $w_t = \tilde{w}_t - \mu$. Note that the LDS will not generate observations with the structure of a one-hot vector shifted by a constant mean, so we cannot use it directly as a generative language model. On the other hand, we can still fit models to training data with this structure, perform posterior inference given observations, assess the likelihood of a corpus, etc. In our experiments, we demonstrate the usefulness of these in a variety of applications.  We have:
\begin{equation}
\label{eq:psi0}
\Psi_0 = \Ex[w_t w_t^\top] = \Ex[\tilde{w}_t \tilde{w}_t^\top] - \mu \mu^\top =  \text{diag}(\mu) - \mu \mu^\top, 
\end{equation}
while at higher lags, 
\begin{equation}
\label{eq:psiK}
\Psi_k = \Ex[w_t w_{t+k}^\top] = \Ex[\tilde{w}_t \tilde{w}_{t+k}^\top] - \mu \mu^\top.
\end{equation}

Approximating these covariances from a length-$T$ corpus:

\begin{equation}
\mu_i = Ex[\tilde{w}_t] = \frac{1}{T}\text{\#}(\text{word } i \text{ appears}),
\end{equation}
where $\text{\#}()$ denotes the count of an event. 
We also have
\begin{align}
\label{eq:psiKb}
& \Ex[\tilde{w}_t \tilde{w}_{t+k}^\top]_{i,j} =\\  \nonumber
&\frac{1}{T} \text{\#}(\text{word } i\text{ appears} \text{ with word } j \; k \text{ positions to the right}).
\end{align}

For real-world data, ~\eqref{eq:psiKb} will be extremely sparse, with the number of nonzeros substantially less than both $V^2$ and the length of the corpus. The fact that~\eqref{eq:psiK} is sparse-minus-low-rank and~\eqref{eq:psi0} is diagonal-minus-low-rank is critical for scaling up the learning algorithms. First of all, we do not instantiate these as $V\times V$ dense matrices, but operate directly on their factorized structure. Second, in Sec.~\ref{sec:full-rank} we show how the structure of~\eqref{eq:psi0} allows us to model full-rank $V\times V$ noise covariance matrices implicitly. Strictly speaking, the number of nonzeros in~\eqref{eq:psiKb} will increase as the corpus size increases, due to heavy-tailed word co-occurence statistics. However, this growth is sublinear in $T$ and can be mitigated by ignoring rare words.   
  
Unfortunately, each $\Psi_k$ is rank-deficient.  Not only is $\Ex[w_t] = 0$, but also the sum of every $w_t$ is zero (because $\tilde{w}_t$ is a one-hot vector and $\mu$ is a vector of word frequencies). Define $\textbf{1}$ to be the length-$V$ vector of ones. Then, our data lives on $\textbf{1}^{\perp}$, the $d-1$ dimensional subspace, orthogonal to $\textbf{1}$. Doing maximum likelihood in $\real^V$ instead of $\textbf{1}^{\perp}$ will lead to a degenerate likelihood function, since the empirical variance in the $\textbf{1}$ direction is 0. However, projecting the data to this subspace breaks the special structure described above, so we instead work in $\real^V$ and perform projections onto $\textbf{1}^{\perp}$ implicitly as-needed. Fortunately, both SSID and EM find $C$ that lies in the column space of the data, so iterations of our  learning algorithm will maintain that $\textbf{1} \notin \text{col(C)}$. In Appendix~\ref{app:KG}, we describe how to handle this rank deficiency when computing the Kalman gain. 

Note that we could have used pretrained type-level embeddings to project our corpus and then train an LDS on low-dimensional dense observations. However, this is vulnerable to the subspace of the type-level embeddings, which are not trained to maximize the likelihod of a sequence model, and thus might not capture proper syntactic and semantic information. We will release the code of our implementation. SSID requires simple scripting on top of a sparse linear algebra library. Our EM implementation consists of small modifications to Martens' public ASOS code.   

\subsection{Scalable Spectral Decomposition}
SSID requires a rank-$h$ SVD of the very large block Hankel matrix $H_r$~\eqref{eq:hankel-matrix}. We employ the randomized approximate SVD algorithm of~\citet{halko2011finding}. To factorize a matrix $X$, this requires repeated multiplication by $X$ and by $X^\top$. All the submatrices in $H_r$  are sparse-minus-low-rank, so we handle the sparse and low-rank terms individually within the multiplication subroutines. 

\subsection{Modeling Full-Rank Noise Covariance}
\label{sec:full-rank}
The noise covariance matrix $D$ is $V \times V$, which is unmanageably large for our application, and thus it is reasonable to employ a spherical $D = dI$ or diagonal $D = \text{diag}(d_1,\ldots,d_V)$ approximation. For our problem, however, we found that these approximations performed poorly. Because of the property $\textbf{1}^\top w_t = 0$,  off-diagonal elements of $D$ are critical for modeling the anti-correlations between coordinates. This would have been captured if we passed $w_t$ through a logistic multinomial link function. However, this prevents simple inference using Kalman filter. To maintain conjugacy, practitioners sometimes employ the quadratic upper bound to a logistic multinomial likelihood introduced in~\citet{bohning1992multinomial}, which  hard-codes the coordinate-wise anticorrelations via $D= \frac{1}{2}\left[I  - \frac{1}{V+1}\mathbf{1}\mathbf{1}^\top \right]$. However,  we found this data-independent estimator performed poorly. 

Instead, we exploit a particular property of the SSID and EM estimators for $D$ in~\eqref{eq:D-SSID} and~\eqref{eq:D-EM}. Namely, both set $D$ to $\Psi_0$ minus a low-rank matrix, and thus $D$ is diagonal-minus-low-rank, due to the structure in~\eqref{eq:psi0}. For the LDS, we mostly seek to manipulate the precision matrix $D^{-1}$. While instantiating this dense $V \times V$ matrix is infeasible, multiplication by $D^{-1}$ and evaluation of $\text{det}(D^{-1})$ can both be done efficiently using the Sherman–Woodbury-Morrison formula (Appendix~\ref{app:MIL}).  In Appendix~\ref{sec:lik}, we also leverage the formula to efficiently evaluate the training likelihood.  These uses of the formula differ from its common usage for LDS, when not using the steady-state assumption and the posterior precision matrix needs to be updated using rank one updates to the covariance. Our technique is particular to fitting indicator-vector data as a multivariate Gaussian. 
 \vspace{-7pt}
\subsection{Whitening}
\label{sec:whiten}
Before applying our learning algorithms, we first whiten the $\Psi$ matrices with the diagonal transformation. 
\begin{equation}
W = \Psi_0^{-\frac{1}{2}} = \text{diag}(\mu_1^{-\frac{1}{2}}, \ldots, \mu_V^{-\frac{1}{2}}). 
\end{equation}
Fitting to $W \Psi_k W^\top$, rather than $\Psi_k$, maintains the data's sparse-minus-low-rank  and diagonal-minus-low-rank structures. Furthermore, EM is unaffected, i.e., applying EM to linearly-transformed data is equivalent to learning on the original data and then transforming post-hoc. 

On the other hand, the SSID output is affected by whitening, since the squared reconstruction loss that SVD implicitly minimizes depends on the coordinate system of the data. We found such whitening crucial for obtaining high-quality initial parameters. Whitening for  SSID, which is recommended by~\citet{van1996subspace}, solves a very similar factorization problem as canonical correlation analysis between words and their contexts, which has been used successfully to learn word embeddings~\citep{dhillon11multiviewcca,dhillon_icml12_tscca} and identify the parameters of class-based language models~\cite{stratosspectral}). 

In Appendix~\ref{app:D-psd} we also provide an algorithm, which relies on whitening, for manually ensuring the $D$ returned by SSID is PSD, without needing to  factorize a $V \times V$ matrix. Such manual correction is unnecessary during EM, since the estimator~\eqref{eq:D-EM} is guaranteed to be PSD. 
\vspace{-7pt}
\section{Embedding Tokens using the LDS}
\label{sec:filtering}
The only data-dependent term in the steady-state filtering and smoothing equations~\eqref{eq:kf1} and~\eqref{eq:kf2} is $K w_t$. Since $w_t$ can take on only $V$ possible values, we precompute these word-type-level  vectors. The computational cost of filtering/smoothing a length $T$ sequence is $O(Th^2)$, which is identical to the cost of inference on a discrete first-order sequence model.~\eqref{eq-ss-gain} is not directly usable to obtain $K$, due to the data's rank-deficiency, and we provide an efficient alternative in Appendix~\ref{app:KG}. This also requires the matrix inversion lemma to avoid instantiating $S^{-1}$ in~\eqref{eq-ss-gain}. 

In our experiments we use the latent space to define features for tokens. However, distances in this space are not well-defined, since the likelihood is invariant to any linear transformation of the latent variables. To place $x_t$ in reasonable coordinates, we compute the empirical posterior covariance $M  = \Ex[\bar{x}\bar{x}^\top]$ on the training data (using ASOS). Then, we whiten $x_t$ using $M^{-\frac{1}{2}}$ and project the result onto the unit sphere. 

\section{Relation to Recurrent Neural Networks}
\label{sec:rnn}
We now highlight the similarity between the parametrization of an RNN architecture commonly used for language modeling and our Kalman filter. This allows us to use our LDS as a novel method for initializing the parameters of a non-linear RNN, which we explore in Sec.~\ref{sec:expts-rnn}. Following~\citep{mikolov2012statistical} we consider the network structure:
\begin{eqnarray}
h_t &=& \sigma(A h_{t-1} + B w_{t-1}) \label{eq:rnn1}\\
w_t &\sim& \text{SoftMax}(C h_t), \label{eq:rnn2}
\end{eqnarray}
Here, we employ the SoftMax transformation of a vector $v$ as $v_i \rightarrow \exp(v_i)/ \sum_k \exp(v_k)$. The coordinate-wise nonlinearity $\sigma(\cdot)$ is, for example, a sigmoid, and the network is initialized with some fixed vector $h_0$. 

Consider the use of the steady-state Kalman filter~\eqref{eq:kf1} as an online predictor, where the mean prediction for $w_t$ is given by $C \hat{x}_{t}$. Then, if  we replace $\sigma$ and $\text{SoftMax}$ with the identity, the Kalman filter and the RNN have the same set of parameters, where we $B$ corresponds to $K$ and $A$ corresponds to $(A - K C A)$. In terms of the state dynamics, the LDS may provide parameters that are reasonable for a nonlinear RNN, since the sigmoid $\sigma$ has a regime for inputs close to zero where it behaves like the identity. A linear approximation of $\text{SoftMax}()$ ignores mutual exclusivity. However, we discuss in Section~\ref{sec:full-rank} that using a full-rank $D$ captures some coordinate-wise anti-correlations. Also,~\eqref{eq:rnn2} does not affect the state evolution in~\eqref{eq:rnn1}. 

A key difference between the LDS and the RNN is that the LDS provides a backwards pass, using Kalman smoothing, where $\bar{x}_t$ depends on words to the right. For RNNs, this would requires separate model~\citep{schuster1997bidirectional}. 

\section{Experiments}

\subsection{LDS Transition Dynamics}

Many popular word embedding methods learn word-to-vector mappings, but do not learn the dynamics of text's evolution in the latent space. Using the specific LDS model we describe in the next section, we employ the transition matrix $A$ to explore properties of these dynamics. Because the state evolution is linear, it can be studied easily using a spectral decomposition. Namely, $A$ converts its left singular vectors into (scaled) right singular vectors. For each vector, we find the words most likely to be generated from this state. Table~\ref{tab:nn} presents these singular vector pairs. We find they reflect interpretable transition dynamics. In all but the last block, the vectors reflect strict state transitions. However, in the last block contains topical terms about food invariant under $A$. Overall, we did not find such salient structure in the parameters estimated using SSID.

\begin{table}
\small
\begin{center}
\begin{tabular}{| c | c |}
\hline
Right Singular Vector & Left Singular Vector\\
\hline
 islamist lebanese israeli &  territories immigrants  sea \\
palestinian british latin & films communities nationals \\
japanese shiite greek & rivals africa clients \\
\hline
chris mike steve &  evans anderson harris\\
 jason tim jeff & robinson smith phillips \\
bobby ian greg & collins murray murphy \\
\hline
singh berlusconi sharon & shares referee suggesting \\
 blair putin abbas &  industries testified insisted \\
netanyahu brown levy &  adding arguing yesterday \\
\hline
tampa colorado minnesota & bay derby division\\
detroit cleveland phoenix & county sox district\\
indiana seattle dallas & sole river valley ballet\\
\hline
 policemen  helicopters soldiers & remained expressed outst\\
suspects demonstrators guards & recommended  remains feels\\
iraqis personnel detainees &  gets resumed sparked \\
\hline
salt chicken pepper & chicken cream pepper \\
chocolate butter cheese & sauce cheese chocolate \\
cream sauce bread &  salt butter bread\\ 
\hline
\end{tabular}
\caption{Words likely to be generated for singular vector pairs of the LDS transition operator. The operator maps right vectors to left, and the pairs are syntactically and semantically coherent. }
\label{tab:nn}
\end{center}
\vspace{-15pt}
\end{table}

\subsection{POS Tagging}
\label{sec:pos}

Unsupervised learning of generative discrete state models for text has been shown to capture part-of-speech (POS) information~\citep{christodoulopoulos2010two}. In response, we assess the ability of the LDS to also capture POS structure. Token embeddings can be used to predict POS in two ways: (1) by applying a local classifier to each token's embedding, or (2) by including each token's embedding as additional features in a lexicalized tagger. For both, we train the tagging model on the Penn Treebank (PTB) train set, which is not included for LDS training. Token embeddings are obtained from Kalman smoothing. We evaluate tagging accuracy on the PTB  test set using the 12 `universal' POS tags~\citep{petrov2011universal} and the original tags. We contrast the LDS with type embeddings from Word2Vec,  trained on the LDS data~\citep{mikolov2013efficient}. 

We fit our LDS using a combination of the APNews, New York Times, and RCV1 newswire corpora, about 1B tokens total. We maintain punctuation and casing of the text, but replace all digits with ``NUM' and all but the most 200k frequent types with ``OOV.''  We employ $r =4$ for SSID, $r = 7$ for EM, and $h = 200$. We add 1000 psuedocounts for each type, by adding $\frac{1000}{T}$ to each coordinate of $\mu$. 

The LDS hyperparameters were selected by maximizing the accuracy of a local classifier on the PTB dev set. This also included when to terminate EM. For Word2Vec, we performed a broad search over hyperparameters, again maximizing for local POS tagging. Our local classifier was a two-layer neural network with 25 hidden units, which outperformed a linear classifier. The best Word2Vec configuration used the CBOW architecture with a window width of 3.  The lexicalized tagger's hyperparameters were also tuned on the PTB dev set.  For the local tagging, we ignored punctuation and few common words types such as ``and'' in  training. Instead, we classified them directly using their majority tag in the training data. 

Overall, we found that the LDS and Word2Vec took about 12 hours to train on a single-core CPU. Since the Word2Vec algorithm is simple and the code is heavily optimized, it performs well, but our learning algorithm would have been substantially faster given a larger training set, since the $\Psi_k$ matrices can be gathered in parallel and the cost of SSID and ASOS is sublinear in the corpus size. In Section~\ref{sec:expts-rnn}, training the LDS is order of magnitude faster than an RNN. 

\begin{table}
\begin{center}
\small
\begin{tabular}{| c | c | c | c | c | c | c | c |}
 \hline 
tags & W2V & SSID &  EM & Lex & Lex & Lex \\
& & &  & & +EM &+W2V \\
\hline 
U & 95.00 & 89.26 & 96.44 & 97.97 & 98.05  & 98.02 \\
\hline
P &  92.58  & 83.00  & 94.30 & 97.28 &  97.32 & 97.35\\
\hline 
\end{tabular}
\caption{POS tagging with universal tags (U) and PTB tags (P).}
\label{tab:pos}
\end{center}
\vspace{-25pt}
\end{table}

Our results are shown in Table~\ref{tab:pos}. Left to right, we compare Word2Vec (W2V), SSID, EM initialized with SSID (EM), our baseline lexicalized tagger (LEX), the lexicalized tagger with extra features from LDS token embeddings (LEX + EM), and the lexicalized tagger with type-level Word2Vec embeddings (LEX + SSID). 

The first 3 columns perform local classification. First, while SSID is crucial for EM initialization, we found it performed poorly on its own. However, EM outperforms Word2Vec substantially. We expect this is because the LDS explicitly maximizes the likelihood of text sequences, and thus it forces token embeddings to capture the transition dynamics of syntax.  All differences are statistically significant at a .05 significance level using the exact binomial test.  In Appendix~\ref{app:init}, we demonstrate the importance of SSID vs. random initialization. The final 3 columns use a carefully-engineered tagger. For universal tags, LDS and Word2Vec both contribute a statistically-significant gain over the baseline (Lex), but their difference is not significant. For PTB tags, we find that Word2Vec achieves a significant gain over LEX, but the LDS does not. We expect that our context-dependent embeddings perform as well as context-independent embeddings since the taggers' features and test-time inference capture non-local interactions.

\subsection{Named Entity Recognition}
In Table~\ref{tab:ner} we consider the effect of unsupervised token features for NER on the Conll 2003 dataset using a lexicalized tagger (Lex).  We use the same LDS and Word2Vec models as in the previous section, and also compare to the Brown clusters used for NER in~\citet{ratinov2009design}.  As before, we find that Word2Vec and LDS provide significant accuracy improvements over the baseline. We expect that the reason the LDS does not outperform Word2Vec is that NER relies mainly on performing local pattern matching, rather than capturing long-range discourse structure. 

\begin{table}[h!]
\begin{center}
\small
\begin{tabular}{| c | c | c | c | c |}
 \hline 
set & Lex & Lex+Brown & Lex+W2V & Lex+LDS \\
\hline 
dev & 93.90 & 93.79 & 94.14  & 94.21 \\
\hline
test &  89.34 & 89.76 & 90.00 & 89.9 \\
\hline 
\end{tabular}
\caption{NER with various unsupervised token features}
\label{tab:ner}
\end{center}
\vspace{-15pt}
\end{table}

\subsection{RNN initialization}
\label{sec:expts-rnn}
\begin{figure}[b]
\vspace{-10pt}
    \centering
    \includegraphics[width=0.65\columnwidth]{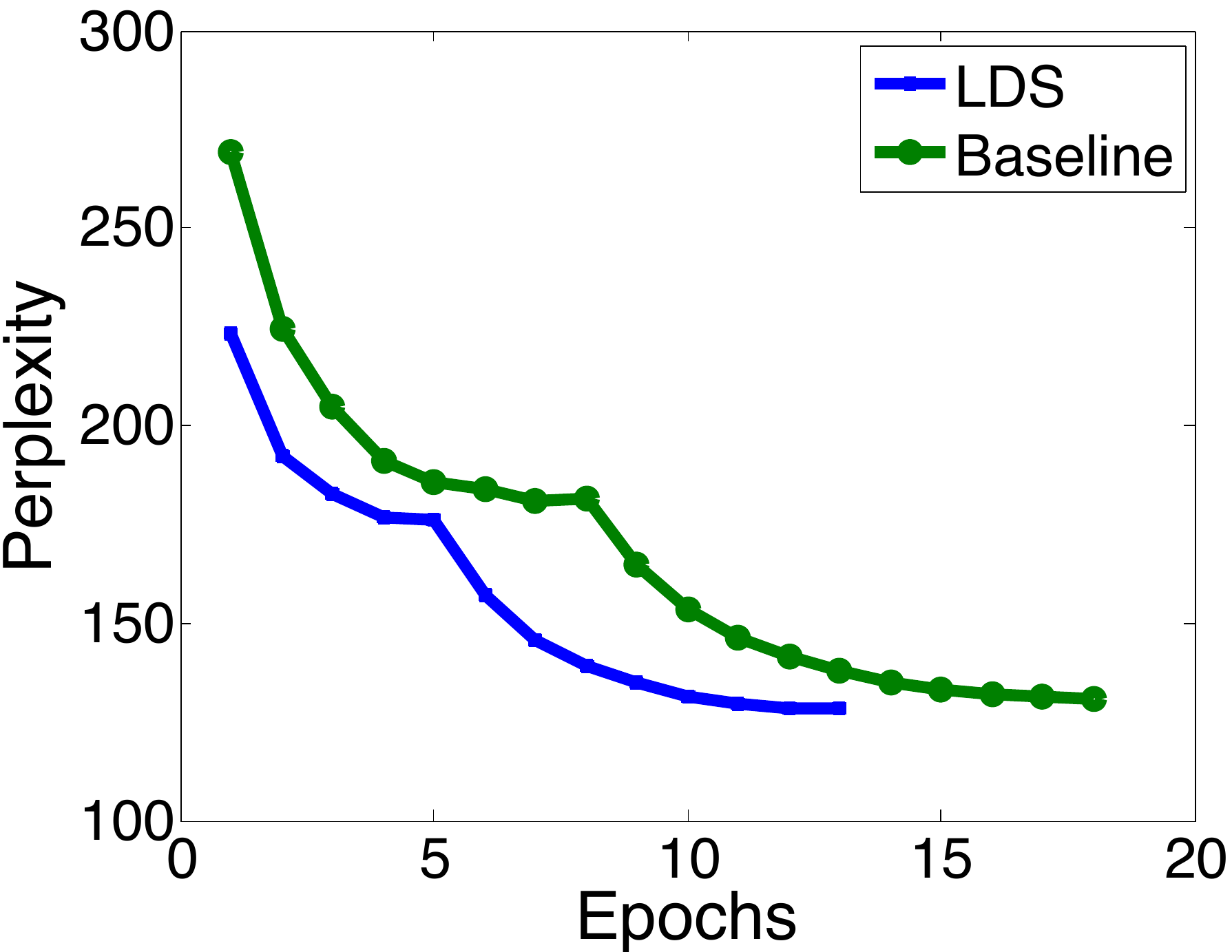}
    \caption{RNN Dev set perplexity vs. \# passes over the training data for baseline initialization vs. LDS initialization.}
    \label{fig:ppCurves}
\end{figure}
As highlighted in Section~\ref{sec:related-work}, RNNs can provide impressive accuracy in various applications. We consider the simple RNN architecture of Sec.~\ref{sec:rnn}, since it permits natural initialization with an LDS and because~\citet{mikolov2014learning} demonstrate that small variants of it can outperform LSTMs as a language model (LM).  Note that the  `context units' of~\citet{mikolov2014learning} could also be learned using our EM procedure, by restricting the parametrization of $A$. We leave exploration of  hierarchical softmax observations~\citep{mnih2009scalable}, and other alternative architectures, for future work. 

We evaluate the usefulness of the LDS for initializing the RNN under two criteria: (1) whether it improves the perplexity of the final model, and (2) whether it leads to faster optimization. A standard dataset for comparing language models is the Penn Treebank (PTB)~\citep{sundermeyer2012lstm,pachitariu2013regularization,mikolov2014learning}. We first train a baseline, obtaining the same test set perplexity as~\citet{mikolov2012statistical}, with 300 hidden dimensions. This initializes parameters randomly, with lengthscales tuned as in~\citet{mikolov2012statistical}. Next, we use the LDS to initialize an RNN. In order to maintain a fair comparison vs. the baseline, we train the LDS on the same PTB data, though in practice one should train it on a substantially larger corpus. 

 We use the popular RNN learning rate schedule where it is constant until performance on held-out data fails to improve, and then it is decreased geometrically until the held-out performance again fails to improve~\citep{mikolov2012statistical}. We tuned the initial value and decay rate. When initializing with the LDS, small learning rates are crucial: otherwise, the optimization jumps far from where it started. 

In Figure~\ref{fig:ppCurves}, we plot perplexity on the dev set vs. the number of training epochs. The time to train the LDS, about 30 minutes, is inconsequential compared to training the RNN (4 days) on a single CPU core. LDS training on the PTB is faster than our experiments above with 1B tokens because we use a small vocabulary and run far fewer EM iterations, in order to prevent overfitting. The RNN baseline converged after 17 training epochs, while using the LDS for initialization allowed it to converge after 12, which amounts to about a day of savings on a single CPU core. Next, in Table~\ref{tab:rnn} we compare the final perplexities on the dev and test sets. We find that initializing with the LDS also provides a better model. 

We found that initializing the RNN with LDS parameters trained using SSID, rather than SSID+EM, performed no better than the baseline. Specifically, the best performance was obtained using a high initial learning rate, which allows gradient descent to ignore the SSID values. We expect this is because the method of moments requires lots of data, and the PTB is small. In a setting where one trains the LDS on a very large corpus, it is possible that SSID is effective.  Overall, we did not explore initializing the RNN using type-level embeddings such as Word2Vec, since it is unclear how to initialize $A$ and how to set $K$ vs. $C$.

\begin{table}
\begin{center}
\small
\begin{tabular}{| c | c | c |}
 \hline 
  & Baseline & LDS\\
\hline
dev & 130.6 & 128.1 \\
\hline
test &  124.0 & 122.8\\
\hline 
\end{tabular}
\caption{Final perplexity for an RNN language model trained using random parameter initialization vs. LDS initialization.}
\label{tab:rnn}
\end{center}
\vspace{-28pt}
\end{table}

\vspace{-7pt}
\section{Conclusion and Future Work}

We have contributed a scalable method for assigning word tokens context-specific low-dimensional representation that capture useful syntactic and semantic structure. Our algorithm requires a single pass over the training data and no painful tuning of learning rates. 

Next, we will extend ASOS to new models and improve initialization for alternative RNN architectures, including hierarchical softmaxes, by leveraging not just the LDS parameters, but also the LDS posterior on the training data. 
\vspace{-10pt}
\section*{Acknowledgments}
This work was partially completed while the first author was an intern at Microsoft Research. We appreciate helpful comments from Brendan O'Connor, Ben Marlin, Andrew McCallum, and Qingqing Huang. 

\bibliography{../sources.bib}

\begin{thebibliography}{43}
\providecommand{\natexlab}[1]{#1}
\providecommand{\url}[1]{\texttt{#1}}
\expandafter\ifx\csname urlstyle\endcsname\relax
  \providecommand{\doi}[1]{doi: #1}\else
  \providecommand{\doi}{doi: \begingroup \urlstyle{rm}\Url}\fi

\bibitem[Anandkumar et~al.(2014)Anandkumar, Ge, Hsu, Kakade, and
  Telgarsky]{anandkumar2014tensor}
Anandkumar, Animashree, Ge, Rong, Hsu, Daniel, Kakade, Sham~M, and Telgarsky,
  Matus.
\newblock Tensor decompositions for learning latent variable models.
\newblock \emph{The Journal of Machine Learning Research}, 15\penalty0
  (1):\penalty0 2773--2832, 2014.

\bibitem[Bansal et~al.(2014)Bansal, Gimpel, and Livescu]{bansal2014tailoring}
Bansal, Mohit, Gimpel, Kevin, and Livescu, Karen.
\newblock Tailoring continuous word representations for dependency parsing.
\newblock In \emph{Proceedings of the Annual Meeting of the Association for
  Computational Linguistics}, 2014.

\bibitem[Bengio et~al.(2006)Bengio, Schwenk, Sen{\'e}cal, Morin, and
  Gauvain]{bengio2006neural}
Bengio, Yoshua, Schwenk, Holger, Sen{\'e}cal, Jean-S{\'e}bastien, Morin,
  Fr{\'e}deric, and Gauvain, Jean-Luc.
\newblock Neural probabilistic language models.
\newblock In \emph{Innovations in Machine Learning}, pp.\  137--186. Springer,
  2006.

\bibitem[B{\"o}hning(1992)]{bohning1992multinomial}
B{\"o}hning, Dankmar.
\newblock Multinomial logistic regression algorithm.
\newblock \emph{Annals of the Institute of Statistical Mathematics},
  44\penalty0 (1):\penalty0 197--200, 1992.

\bibitem[Brown et~al.(1992)Brown, Desouza, Mercer, Pietra, and
  Lai]{brown1992class}
Brown, Peter~F, Desouza, Peter~V, Mercer, Robert~L, Pietra, Vincent J~Della,
  and Lai, Jenifer~C.
\newblock Class-based n-gram models of natural language.
\newblock \emph{Computational linguistics}, 18\penalty0 (4):\penalty0 467--479,
  1992.

\bibitem[Chelba \& Jelinek(2000)Chelba and Jelinek]{chelba2000structured}
Chelba, Ciprian and Jelinek, Frederick.
\newblock Structured language modeling.
\newblock \emph{Computer Speech \& Language}, 14\penalty0 (4):\penalty0
  283--332, 2000.

\bibitem[Christodoulopoulos et~al.(2010)Christodoulopoulos, Goldwater, and
  Steedman]{christodoulopoulos2010two}
Christodoulopoulos, Christos, Goldwater, Sharon, and Steedman, Mark.
\newblock Two decades of unsupervised pos induction: How far have we come?
\newblock In \emph{Proceedings of the 2010 Conference on Empirical Methods in
  Natural Language Processing}, pp.\  575--584. Association for Computational
  Linguistics, 2010.

\bibitem[Collobert et~al.(2011)Collobert, Weston, Bottou, Karlen, Kavukcuoglu,
  and Kuksa]{collobert2011natural}
Collobert, Ronan, Weston, Jason, Bottou, L{\'e}on, Karlen, Michael,
  Kavukcuoglu, Koray, and Kuksa, Pavel.
\newblock Natural language processing (almost) from scratch.
\newblock \emph{The Journal of Machine Learning Research}, 12:\penalty0
  2493--2537, 2011.

\bibitem[Dhillon et~al.(2012)Dhillon, Rodu, Foster, and
  Ungar]{dhillon_icml12_tscca}
Dhillon, Paramveer, Rodu, Jordan, Foster, Dean, and Ungar, Lyle.
\newblock Two step cca: A new spectral method for estimating vector models of
  words.
\newblock In \emph{Proceedings of the 29th International Conference on Machine
  learning}, ICML'12, 2012.

\bibitem[Dhillon et~al.(2011)Dhillon, Foster, and Ungar]{dhillon11multiviewcca}
Dhillon, Paramveer~S., Foster, Dean, and Ungar, Lyle.
\newblock Multi-view learning of word embeddings via cca.
\newblock In \emph{Advances in Neural Information Processing Systems (NIPS)},
  volume~24, 2011.

\bibitem[Ghahramani \& Hinton(1996)Ghahramani and
  Hinton]{ghahramani1996parameter}
Ghahramani, Zoubin and Hinton, Geoffrey~E.
\newblock Parameter estimation for linear dynamical systems.
\newblock Technical report, Technical Report CRG-TR-96-2, University of
  Totronto, Dept. of Computer Science, 1996.

\bibitem[Ghahramani \& Roweis(1999)Ghahramani and
  Roweis]{ghahramani1999learning}
Ghahramani, Zoubin and Roweis, Sam~T.
\newblock Learning nonlinear dynamical systems using an em algorithm.
\newblock \emph{Advances in neural information processing systems}, pp.\
  431--437, 1999.

\bibitem[Halko et~al.(2011)Halko, Martinsson, and Tropp]{halko2011finding}
Halko, Nathan, Martinsson, Per-Gunnar, and Tropp, Joel~A.
\newblock Finding structure with randomness: Probabilistic algorithms for
  constructing approximate matrix decompositions.
\newblock \emph{SIAM review}, 53\penalty0 (2):\penalty0 217--288, 2011.

\bibitem[Huang et~al.(2012)Huang, Socher, Manning, and Ng]{huang2012improving}
Huang, Eric~H, Socher, Richard, Manning, Christopher~D, and Ng, Andrew~Y.
\newblock Improving word representations via global context and multiple word
  prototypes.
\newblock In \emph{Proceedings of the 50th Annual Meeting of the Association
  for Computational Linguistics: Long Papers-Volume 1}, pp.\  873--882.
  Association for Computational Linguistics, 2012.

\bibitem[Kalman(1960)]{kalman}
Kalman, Rudolph~Emil.
\newblock A new approach to linear filtering and prediction problems.
\newblock \emph{Transactions of the ASME--Journal of Basic Engineering},
  82\penalty0 (Series D):\penalty0 35--45, 1960.

\bibitem[Le~Cam(1974)]{le1974notes}
Le~Cam, Lucien~Marie.
\newblock \emph{Notes on asymptotic methods in statistical decision theory},
  volume~1.
\newblock Centre de Recherches Math{\'e}matiques, Universit{\'e} de
  Montr{\'e}al, 1974.

\bibitem[Levy \& Goldberg(2014)Levy and Goldberg]{levy2014neural}
Levy, Omer and Goldberg, Yoav.
\newblock Neural word embedding as implicit matrix factorization.
\newblock In \emph{Advances in Neural Information Processing Systems}, pp.\
  2177--2185, 2014.

\bibitem[Martens(2010)]{martens2010learning}
Martens, James.
\newblock Learning the linear dynamical system with asos.
\newblock In \emph{Proceedings of the 27th International Conference on Machine
  Learning (ICML-10)}, pp.\  743--750, 2010.

\bibitem[Mikolov(2012)]{mikolov2012statistical}
Mikolov, Tom{\'a}{\v{s}}.
\newblock \emph{Statistical language models based on neural networks}.
\newblock PhD thesis, Ph. D. thesis, Brno University of Technology, 2012.

\bibitem[Mikolov et~al.(2013)Mikolov, Sutskever, Chen, Corrado, and
  Dean]{mikolov2013efficient}
Mikolov, Tomas, Sutskever, Ilya, Chen, Kai, Corrado, Greg~S, and Dean, Jeff.
\newblock Distributed representations of words and phrases and their
  compositionality.
\newblock In \emph{Advances in Neural Information Processing Systems}, pp.\
  3111--3119, 2013.

\bibitem[Mikolov et~al.(2015)Mikolov, Joulin, Chopra, Mathieu, and
  Ranzato]{mikolov2014learning}
Mikolov, Tomas, Joulin, Armand, Chopra, Sumit, Mathieu, Michael, and Ranzato,
  Marc'Aurelio.
\newblock Learning longer memory in recurrent neural networks.
\newblock \emph{International Conference on Learning Representations}, 2015.

\bibitem[Mnih \& Hinton(2009)Mnih and Hinton]{mnih2009scalable}
Mnih, Andriy and Hinton, Geoffrey~E.
\newblock A scalable hierarchical distributed language model.
\newblock In \emph{Advances in neural information processing systems}, pp.\
  1081--1088, 2009.

\bibitem[Neelakantan et~al.(2014)Neelakantan, Shankar, Passos, and
  McCallum]{neelakantan2014efficient}
Neelakantan, Arvind, Shankar, Jeevan, Passos, Alexandre, and McCallum, Andrew.
\newblock Efficient nonparametric estimation of multiple embeddings per word in
  vector space.
\newblock In \emph{Proceedings of EMNLP}, 2014.

\bibitem[Pachitariu \& Sahani(2013)Pachitariu and
  Sahani]{pachitariu2013regularization}
Pachitariu, Marius and Sahani, Maneesh.
\newblock Regularization and nonlinearities for neural language models: when
  are they needed?
\newblock \emph{arXiv preprint arXiv:1301.5650}, 2013.

\bibitem[Pasa \& Sperduti(2014)Pasa and Sperduti]{pasa2014pre}
Pasa, Luca and Sperduti, Alessandro.
\newblock Pre-training of recurrent neural networks via linear autoencoders.
\newblock In \emph{Advances in Neural Information Processing Systems}, pp.\
  3572--3580, 2014.

\bibitem[Passos et~al.(2014)Passos, Kumar, and
  McCallum]{passos-kumar-mccallum:2014:W14-16}
Passos, Alexandre, Kumar, Vineet, and McCallum, Andrew.
\newblock Lexicon infused phrase embeddings for named entity resolution.
\newblock In \emph{Proceedings of the Eighteenth Conference on Computational
  Natural Language Learning}, 2014.

\bibitem[Pennington et~al.(2014)Pennington, Socher, and
  Manning]{pennington2014glove}
Pennington, Jeffrey, Socher, Richard, and Manning, Christopher~D.
\newblock Glove: Global vectors for word representation.
\newblock \emph{Proceedings of the Empiricial Methods in Natural Language
  Processing (EMNLP 2014)}, 12, 2014.

\bibitem[Petrov et~al.(2011)Petrov, Das, and McDonald]{petrov2011universal}
Petrov, Slav, Das, Dipanjan, and McDonald, Ryan.
\newblock A universal part-of-speech tagset.
\newblock \emph{arXiv preprint arXiv:1104.2086}, 2011.

\bibitem[Press et~al.(1987)Press, Flannery, Teukolsky, and
  Vetterling]{press1987numerical}
Press, William~H, Flannery, Brian~P, Teukolsky, Saul~A, and Vetterling,
  William~T.
\newblock \emph{Numerical Recipes: The art of scientific computing}, volume~2.
\newblock Cambridge University Press London, 1987.

\bibitem[Ratinov \& Roth(2009)Ratinov and Roth]{ratinov2009design}
Ratinov, Lev and Roth, Dan.
\newblock Design challenges and misconceptions in named entity recognition.
\newblock In \emph{Proceedings of the Thirteenth Conference on Computational
  Natural Language Learning}, pp.\  147--155. Association for Computational
  Linguistics, 2009.

\bibitem[Roweis \& Ghahramani(1999)Roweis and Ghahramani]{roweis1999unifying}
Roweis, Sam and Ghahramani, Zoubin.
\newblock A unifying review of linear gaussian models.
\newblock \emph{Neural computation}, 11\penalty0 (2):\penalty0 305--345, 1999.

\bibitem[Rugh(1996)]{rugh1996linear}
Rugh, Wilson~J.
\newblock \emph{Linear system theory}, volume~2.
\newblock prentice hall Upper Saddle River, NJ, 1996.

\bibitem[Saxe et~al.(2014)Saxe, McClelland, and Ganguli]{saxe2013exact}
Saxe, Andrew~M, McClelland, James~L, and Ganguli, Surya.
\newblock Exact solutions to the nonlinear dynamics of learning in deep linear
  neural networks.
\newblock \emph{International Conference on Learning Representations}, 2014.

\bibitem[Schuster \& Paliwal(1997)Schuster and
  Paliwal]{schuster1997bidirectional}
Schuster, Mike and Paliwal, Kuldip~K.
\newblock Bidirectional recurrent neural networks.
\newblock \emph{Signal Processing, IEEE Transactions on}, 45\penalty0
  (11):\penalty0 2673--2681, 1997.

\bibitem[Smith et~al.(1999)Smith, de~Freitas, Robinson, and
  Niranjan]{smith2000}
Smith, Gavin, de~Freitas, Joao~FG, Robinson, Tony, and Niranjan, Mahesan.
\newblock Speech modelling using subspace and em techniques.
\newblock \emph{Advances in neural information processing systems}, pp.\
  431--437, 1999.

\bibitem[Socher et~al.(2013)Socher, Chen, Manning, and Ng]{socher2013reasoning}
Socher, Richard, Chen, Danqi, Manning, Christopher~D, and Ng, Andrew.
\newblock Reasoning with neural tensor networks for knowledge base completion.
\newblock In \emph{Advances in Neural Information Processing Systems}, pp.\
  926--934, 2013.

\bibitem[Stratos et~al.(2014)Stratos, Kim, Collins, and Hsu]{stratosspectral}
Stratos, Karl, Kim, Do-kyum, Collins, Michael, and Hsu, Daniel.
\newblock A spectral algorithm for learning class-based n-gram models of
  natural language.
\newblock In \emph{Uncertainty in Artificial Intelligence (UAI)}, 2014.

\bibitem[Sundermeyer et~al.(2012)Sundermeyer, Schl{\"u}ter, and
  Ney]{sundermeyer2012lstm}
Sundermeyer, Martin, Schl{\"u}ter, Ralf, and Ney, Hermann.
\newblock Lstm neural networks for language modeling.
\newblock In \emph{INTERSPEECH}, 2012.

\bibitem[Sutskever et~al.(2014)Sutskever, Vinyals, and
  Le]{sutskever2014sequence}
Sutskever, Ilya, Vinyals, Oriol, and Le, Quoc~VV.
\newblock Sequence to sequence learning with neural networks.
\newblock In \emph{Advances in Neural Information Processing Systems}, pp.\
  3104--3112, 2014.

\bibitem[Turian et~al.(2010)Turian, Ratinov, and Bengio]{turian2010word}
Turian, Joseph, Ratinov, Lev, and Bengio, Yoshua.
\newblock Word representations: a simple and general method for semi-supervised
  learning.
\newblock In \emph{Proceedings of the 48th Annual Meeting of the Association
  for Computational Linguistics}, pp.\  384--394. Association for Computational
  Linguistics, 2010.

\bibitem[Van~Overschee \& De~Moor(1996)Van~Overschee and
  De~Moor]{van1996subspace}
Van~Overschee, Peter and De~Moor, B.
\newblock Subspace identification for linear systems: Theory, implementation.
\newblock \emph{Methods}, 1996.

\bibitem[Van~Overschee \& De~Moor(1994)Van~Overschee and De~Moor]{van1994n4sid}
Van~Overschee, Peter and De~Moor, Bart.
\newblock N4sid: Subspace algorithms for the identification of combined
  deterministic-stochastic systems.
\newblock \emph{Automatica}, 30\penalty0 (1):\penalty0 75--93, 1994.

\bibitem[Vinyals et~al.(2014)Vinyals, Kaiser, Koo, Petrov, Sutskever, and
  Hinton]{vinyals2014grammar}
Vinyals, Oriol, Kaiser, Lukasz, Koo, Terry, Petrov, Slav, Sutskever, Ilya, and
  Hinton, Geoffrey.
\newblock Grammar as a foreign language.
\newblock \emph{arXiv preprint arXiv:1412.7449}, 2014.

\end{thebibliography}
\bibliographystyle{icml2015}

\newpage
\onecolumn
\appendix
\begin{center}
\Large{Supplementary Material}
\end{center}

\section{Scaling UP LDS Learning to Text}

As discussed in Section~\ref{sec:whiten},  we whiten our data using
\begin{equation}
W = \Psi_0^{-\frac{1}{2}} = \text{diag}(\mu_1^{-\frac{1}{2}}, \ldots, \mu_V^{-\frac{1}{2}}). 
\end{equation}
Besides improving the empirical performance of SSID, working in the whitened coordinate system also simplifies various details used in Section~\ref{sec:text-lds} when scaling up LDS learning for text. Under this transformation, we have $\Psi_0 = \text{diag}(\mu) - \mu\mu^\top$. This simplifies various steps because our estimators ~\eqref{eq:D-SSID} and~\eqref{eq:D-EM} are of the form $I - \text{[low rank matrix]}$, rather than  $\text{diag}(\mu) - \text{[low rank matrix]}$. In the whitened coordinates, the data are orthogonal to $\muHalf$, rather than $\textbf{1}$. 

\subsection{Recovering PSD $D$ in SSID }
\label{app:D-psd}
While SSID is consistent, for finite data the procedure is not guaranteed to yield a positive semidefinite (PSD) estimate for $D$, which is required because it is a covariance matrix. In our particular case, the $D$ we seek will be singular on the span of $\muHalf$, but Subspace ID will still not guarantee that $D$ will be PSD on $\muHalf^{\perp}$. 

This is critical because if $D$ is not PSD on this subspace, then we can not define a valid Kalman filtering procedure for the model (see Sec.~\ref{app:KG}). However, due to the structure of our data distribution, $D$ can easily be fixed post-hoc.

From~\eqref{eq:D-SSID} we have the estimator 
\begin{align}
D &= I - \muHalf\muHalf^\top  -  C \Sigma_{1}C^\top 
\end{align}

Next, define
$D_\alpha= I- \muHalf\muHalf^\top  -  (1 - \alpha) C \Sigma_{1}C^\top$
and define the PSD estimator $D^{\prime } = D_{\alpha_0}$, where $\alpha_0$ is the minimal value such that $D_\alpha$ is PSD on $\muHalf^{\perp}$. We next show how to find $\alpha_0$. 

We have that  $D_\alpha$  is PSD on ${\boldsymbol{\mu}^{\frac{1}{2}}}^\perp$ iff the maximum eigenvalue of $(1 - \alpha) C \Sigma_{1}C^\top $ is less than 1. This is because $ \muHalf$ is a unit vector and we can ignore any cross terms between $\muHalf\muHalf^\top$ and $ (1 - \alpha) C \Sigma_{1}C^\top$ because $\text{col}(C) = \muHalf^\perp$, which is true because the data lies in this subspace. Therefore we can find $\alpha_0$ using the following procedure:
 \begin{enumerate}
\item Find $s_0$, the maximal eigenvalue of $C \Sigma_{1}C^\top$, using power iteration. This can be done efficiently by keep $C \Sigma_{1}C^\top$ in its factorized form and not instantiating a $V \times V$ matrix. 
\item If $s_0 <1$, set $\alpha_0 = 0$. Otherwise, set $\alpha_0 = \frac{s_0-1}{s_0}$.
 \end{enumerate}

\subsection{Efficiently Computing the Kalman Gain Matrix}
\label{app:KG}
Next, recall our expression~\eqref{eq-ss-gain} for the steady state Kalman gain $K = \Sigma_1 C^\top S^{-1}_{ss}$, which comes from solving the system 
\begin{equation}
KS_{ss} = \Sigma_1 C^\top, \label{eq:kss-eq}
\end{equation} 
where
\begin{align}
S_{ss} &= C \Sigma_1  C^\top + D 
\end{align}

Furthermore, note that both of our estimators for $D$,~\eqref{eq:D-SSID} and~\eqref{eq:D-EM}, maintain the property that $\muHalf$ is an eigenvector of eigenvalue 0 for $D$.

Since $\muHalf$ is also orthogonal to $\text{col}(C)$, we have that $\muHalf \notin \text{Col}(S_{ss})$. Therefore, we cannot use ~\eqref{eq-ss-gain} directly because $S_{ss}$ is not invertible along this direction. However, we can still solve~\eqref{eq:kss-eq} as $K = \Sigma_1 C^\top S^{+}_{ss}$. This pseudoinverse can be characterized as:
\begin{align}
S^{+}_{ss} &=  \left[\text{inversion of } S_{ss}\text{ within }\text{col}(S_{ss})\right]\left[\text{projection onto } \text{col}(S_{ss})\right]
\end{align}

Furthermore, note that both estimators for D have the form that \begin{align}
D &= \Psi_0 - (\text{PSD, low rank, and } \perp \muHalf) \\
&= I  - \muHalf\muHalf^\top - (\text{PSD, low rank and } \perp \muHalf) \\
&\defeq I  - \muHalf\muHalf^\top - L \label{eq:L}
\end{align}

Therefore, it remains to define the pseudoinverse of 
\begin{equation}
 S_{ss}  = I  - \mu^{\frac{1}{2}}{\mu^{\frac{1}{2}}}^\top +C (\Sigma_1-M)  C^\top).\label{eq:ss1}
\end{equation}

Furthermore, since $\text{col}(L) = \text{col}(C) = \muHalf^\perp$, we can define $L = C MC^\top$ for some positive definite $M$, so we consider

\begin{equation}
 S_{ss}  = I  - \mu^{\frac{1}{2}}{\mu^{\frac{1}{2}}}^\top +  C (\Sigma_1-M)  C^\top).\label{eq:ss}
\end{equation}

Observe that
\begin{equation}
 (I  + C (\Sigma_1-M)  C^\top)^{-1} \label{eq:ss-inv}
\end{equation}
is a valid inverse for $S_{ss}$ on $\muHalf^{\perp}$. This follows from the orthogonality of $\muHalf$ and $\text{col}(C)$, so we can effectively ignore the $\muHalf$ term in~\eqref{eq:ss} when inverting it on  $\muHalf^{\perp}$.

Therefore, we employ
\begin{equation}
(S_{ss} )^+  = (I  +  C (\Sigma_1-M)  C^\top)^{-1} (I - \mu^{\frac{1}{2}}{\mu^{\frac{1}{2}}}), \label{eq:pi}
\end{equation}
where the right term is an orthogonal projection onto $\muHalf^{\perp}$. 

The term in the inverse~\eqref{eq:pi} is diagonal-plus-low-rank and can be manipulated efficiently using the matrix inversion lemma formula~\eqref{eq:swm}:
\begin{equation}
 (I  + C (\Sigma_1-M)  C^\top)^{-1} = I - C((\Sigma_1-M)^{-1}  + C'C)^{-1}C^\top. \label{eq:ss-swm}
\end{equation}

Therefore we can obtain $K$ without instantiating an intermediate matrix of size $V \times V$.

Recall the filtering equation~\eqref{eq:kf1}:
$$ \hat{x}^{t}_t = (A - KC A)\hat{x}^{t-1}_{t-1} + K w_t.  $$

We seek to avoid any $O(V)$ (or worse) computation at test time when filtering. First of all, we can precompute $(A - KC A)$. For the second term, there are only $V$ possible values for the unwhitened input $w_t =\tilde{w}_t - \mu$, so we would like to precompute  $K W(\tilde{w}_t - \mu)$ for every possible value that the indicator $\tilde{w}_t$ can take on. Let $\tilde{w}_t  = e_i$, we have:
\begin{align}
K W(\tilde{w}_t - \mu) &= \Sigma_1 C^\top S_{ss}^+W(e_i - \mu)\\
 &= \Sigma_1 C^\top (I  + C (\Sigma_1-M)  C^\top)^{-1} (I - \mu^{\frac{1}{2}}{\mu^{\frac{1}{2}}}^\top) (W e_i -  \mu^{\frac{1}{2}})\\
&= \Sigma_1 C^\top (I  + C (\Sigma_1-M)  C^\top)^{-1} (I - \mu^{\frac{1}{2}}{\mu^{\frac{1}{2}}}^\top) W e_i\\
&= \Sigma_1 C^\top (I  + C (\Sigma_1-M)  C^\top)^{-1}W   e_i\\
&= \left[\Sigma_1 C^\top (I  + C (\Sigma_1-M)  C^\top)^{-1}W\right]_i,\\
\end{align}
In the final line, the subscript $i$ denotes the $i$th column of a matrix.

\subsection{Likelihood Computation}
\label{sec:lik}
$S_{ss}$ is also used when computing the log-likelihood of input data $(w_1, \ldots,w_T)$:

\begin{equation}
LL = -TV\log(2\pi) - \frac{1}{2}\log\det(S_{ss}) + \sum_{t=1}^\top (w_t^{pred} - w_t)^\top S^{-1}_{ss}(w_t^{pred} - w_t).\label{eq:LL}
\end{equation}
Here, $w_t^{pred} = C A \hat{x}_t$, where $\hat{x}_t$ is the posterior mean for $x_t$ given observations $w_{1:(t-1)}$. $S_{ss}$ is only invertible along $\muHalf^\perp$, but $(w_t^{pred} - w_t)$ varies only on this subspace, so we can effectively ignore the zero-variance direction $\muHalf$. Therefore, we just use~\eqref{eq:ss-inv} as $S_{ss}^{-1}$ in~\eqref{eq:LL}. 

For the data-dependent term in our likelihood, we have:
\begin{align}
&-\frac{1}{2}\sum_{t=1}^\top (w_t^{pred} - w_t)^\top S^{-1}_{ss}(w_t^{pred} - w_t) \\
&=\frac{-1}{2} tr\left(S^{-1}_{ss} \Ex_t[(w_t^{pred} - w_t)(w_t^{pred} - w_t)^\top] \right)\\
&= \frac{-1}{2} tr\left(S^{-1}_{ss} \Ex_t[(w_t - C A \hat{x}_t)(w_t - C A \hat{x}_t)^\top] \right)\\
&=\frac{-1}{2}\left( tr\left(S^{-1}_{ss} \Ex_t[ w_t w_t^\top]\right) -2  tr\left(S^{-1}_{ss} \Ex_t[ w_t \hat{x}_t^\top] A^\top C^\top\right) + tr\left(S^{-1}_{ss} C A \Ex_t[ \hat{x}_t \hat{x}_t^\top] A^\top C^\top \right)\right)\\
&=\frac{-1}{2}\left( tr\left(S^{-1}_{ss} I\right) -2  tr\left(S^{-1}_{ss} \Ex_t[ w_t \hat{x}_t^\top] A^\top C^\top\right) + tr\left(S^{-1}_{ss} C A \Ex_t[ \hat{x}_t \hat{x}_t^\top] A^\top C^\top \right)\right)
\end{align}
Note that the $\Ex_t[ \hat{x}_t \hat{x}_t^\top] $ term above is different from $\Sigma_1$, since the former is from the posterior distribution given the input data and $\Sigma_1$ is from the prior. 

The first term can be computed using~\eqref{eq:swm-trace}. The latter two terms are of the form $tr\left(S^{-1}_{ss} ZW^\top\right)$, where $Z$ and $W$ are both $V \times k$, so we can invoke~\eqref{eq:swm-prod-trace}. For the $\log \det(S_{ss}) $ term, we consider $S_{ss}$ only on $\muHalf^\perp$, so we compute $-\log \det(S^{-1}_{ss})$, where $S^{-1}_{ss}$ comes from~\eqref{eq:ss-inv} and we employ the formula~\eqref{eq:swm-logdet}.

\section{Background}
\subsection{Non-Steady-State Kalman Filtering and Smoothing}
\label{app:filter}
We will use $\hat{x}^\tau_t$ and $S^\tau_t$ for the mean and variance under the posterior for $x_t$ given $w_{1:\tau}$. We will use $\bar{x}_t$ and $S_t^T$ when considering the posterior for $x_t$ given all the data $w_{1:T}$. The following are the forward `filtering' steps~\citep{kalman,ghahramani1996parameter}: 

\begin{align}
\hat{x}^{t-1}_t &= A \hat{x}^{t-1}_{t-1}\\
S^{t-1}_t &= A S_{t-1}^{t-1} A^\top + Q\\
K_t &= S_t^{t-1} C' (C S_{t-1}^{t-1} C^\top + D)^{-1} \label{eq:exact-gain}\\
\hat{x}^{t}_t &= \hat{x}^{t}_{t-1} + K_t (w_t - C \hat{x}^{t-1}_t)\\
S^{t-1}_t &= S^{t-1}_t - K_t C S^{t-1}_t
\end{align}

Next, we have the backwards `smoothing' steps:
\begin{align}
J_{t-1} &= S^{t-1}_{t-1} A' (S^{t-1}_t)^{-1} \label{eq:smoothing}\\
\bar{x}_{t-1} &= \hat{x}_{t-1}^{t-1} + J_{t-1}(\bar{x}_t^T - A \hat{x}_{t-1}^{t-1})\\
S_{t-1}^T &= S_{t-1}^{t-1} + J_{t-1}(S_t^\top - S_t^{t-1})J^T_{t-1}
\end{align}

Note that the updates for the variances $S$ are data-independent and just depend on the parameters of the model. They will converge quickly to time-independent `steady state' quantities.

\subsection{Matrix Inversion Lemma}
\label{app:MIL}
Following~\citet{press1987numerical}, we have
\begin{equation}
(A + U S V^\top)^{-1} = A^{-1} - A^{-1}U(S^{-1} + V^\top A^{-1}U )^{-1}V^\top A^{-1} \label{eq:swm}
\end{equation}
and the related expression for determinants:
\begin{equation}
\det(A + USV^\top) = \det(S)\det(A)\det(S^{-1} + V^\top A^{-1}U). \label{eq:swm-det}
\end{equation}
i.e.
\begin{equation}
\log \det(A + USV^\top) = \log \det(S) + \log\det(A) + \log\det(S^{-1} + V^\top A^{-1}U). \label{eq:swm-logdet}
\end{equation}

Expression~\eqref{eq:swm} is useful if we already have an inverse for $A$ and want to efficiently compute the inverse of a low-rank perturbation of $A$. It is also useful in order to be able to do linear algebra using $(A + U S V^\top)^{-1}$ without actually instantiating a $V \times V$ matrix, which can be unmanageable in terms of both time and space  for large $V$. For example, let $M$ be an $ V \times m$ matrix with $m << V$, then we can compute $M (A + U S V^\top)^{-1}$ using~\eqref{eq:swm} by carefully placing our parentheses such that no $V \times V$ matrix is required. In our application, $A$ is diagonal, so computing its inverse is trivial. Also, note that~\eqref{eq:swm} can be used recursively, if $A$ is defined as another sum of an easily invertible matrix and a low rank matrix. 

Along these lines, here are a few additional useful identities that follow from~\eqref{eq:swm} for quantities that can be computed without $V^2$ time or storage. Here, we assume that both $A^{-1}$ and $tr(A^{-1})$ can be computed inexpensively (e.g., $A$ is diagonal). 

For any product $XY^\top$, where $X$ and $Y$ are $V \times k$ matrices, note that we can compute $tr(XY^T)$ in $O(Vk)$ time as 
\begin{equation}
tr(XY^T) = \sum_i \sum_j X_{ij}Y_{ij}.\label{eq:xy-trace}
\end{equation}
We can use this to compute the trace of the inverse of a matrix implicitly defined via the matrix inversion lemma:
\begin{align}
tr\left[(A + U S V^\top)^{-1}\right] = tr(A^{-1}) - tr\left[\underbrace{A^{-1}U(S^{-1} + V^\top A^{-1}U )^{-1}}_{X}\underbrace{V^\top A^{-1}}_{Y^\top} \right].\label{eq:swm-trace}
\end{align}

More generally, Let $Z$ and $W$ be $V \times k$ matrices, then we compute
\begin{align}
tr \left[ (A + U S V^\top)^{-1}   ZW^\top\right] &= tr(\underbrace{A^{-1}Z}_{X}\underbrace{W^\top}_{Y^\top}) - tr\left[\underbrace{A^{-1}U(S^{-1} + V^\top A^{-1}U )^{-1}}_{X}\underbrace{V^\top A^{-1}ZW^\top}_{Y^\top} \right]\label{eq:swm-prod-trace}
\end{align}
We use~\eqref{eq:swm-prod-trace} when computing the Likelihood in Section~\ref{sec:lik}. 
%

\section{SSID Initialization vs. Random Initialization}
\label{app:init}
\begin{figure}[h!]
    \centering
    \includegraphics[width=0.5\columnwidth]{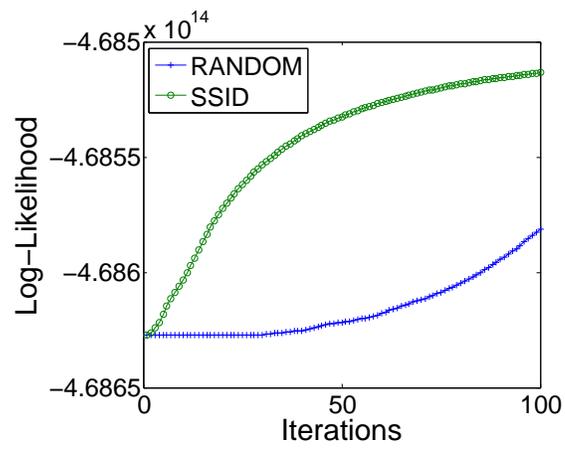}
    \caption{EM Log-Likelihood vs. training iterations for random initialization and SSID initialization.}
    \label{fig:llCurves}
\end{figure}

In Figure~\ref{fig:llCurves}, we contrast the progress of EM, in terms of the log-likelihood of the training data,  when initializing with SSID vs. initializing randomly (Random). Note that the initial values of SSID and Random are nearly identical. This is due to model mispecification, and the fact that we chose the lengthscales of the random parameters post-hoc, by looking at the lengthscales of the SSID parameters. Over the course of 100 EM iterations, the model initialized with SSID climbs quickly and begins leveling out, whereas it takes a long time for the Random model to begin climbing at all. We truncate at 100 EM iterations, since we actually use the SSID-initialized model after the 50th iteration. After that, we find that local POS tagging accuracy diminished.

\end{document}